\documentclass[letterpaper, 10 pt, conference]{ieeeconf}  

\IEEEoverridecommandlockouts                              

\usepackage{cite}
\usepackage{amsmath,amssymb,amsfonts}
\usepackage{algpseudocode}
\usepackage{algorithm}
\usepackage{graphicx}
\usepackage{subcaption}
\usepackage{textcomp}
\usepackage{xcolor}
\usepackage{bbm}
\usepackage{booktabs}
\usepackage[capitalize]{cleveref}
\usepackage{stfloats}

\usepackage{pgfplots}
\usepackage{pgfplotstable}

\usepackage{amsthm}

\pgfplotsset{compat=1.18}
\usepgfplotslibrary{fillbetween}


\newcommand{\needed}[1]{[{\color{blue}citation needed}] }

\newcommand{\statespace}{\mathcal{X}}
\newcommand{\actionspace}{\mathcal{U}}

\newcommand{\unsafe}{\statespace_\text{unsafe}}
\newcommand{\safe}{\statespace_\text{safe}}
\newcommand{\dyn}{f_{\text{dyn}}}

\algnewcommand\algorithmicinput{\textbf{Input:}}
\algnewcommand\Input{\item[\algorithmicinput]}
\algnewcommand\algorithmicoutput{\textbf{Output:}}
\algnewcommand\Output{\item[\algorithmicoutput]}

\newtheorem{theorem}{Theorem}[section]

\newtheorem{definition}[theorem]{Definition}

\crefname{section}{Section}{Sections}
\crefname{theorem}{Theorem}{Theorems}
\crefname{lemma}{Lemma}{Lemmas}
\crefname{table}{Table}{Tables}
\crefformat{equation}{(#2#1#3)}
\crefname{algocf}{Algorithm}{Algorithms}
\Crefname{algocf}{Algorithm}{Algorithms}
\crefname{ALC@unique}{Line}{Lines}

\usetikzlibrary{arrows.meta, positioning, fit, backgrounds, calc}

\definecolor{clrBaseline}{RGB}{31,119,180}    
\definecolor{clrRAC}     {RGB}{44,160,44}     
\definecolor{clrLNT}     {RGB}{255,127,14}    
\definecolor{clrOracle}  {RGB}{140,86,75}     
\definecolor{clrSAVMPC}  {RGB}{214,39,40}     

\definecolor{colEnv}{RGB}{80,80,80}       
\definecolor{colPolicy}{RGB}{45,110,65}   
\definecolor{colInner}{RGB}{90,90,110}    
\definecolor{colSAV}{RGB}{35,70,130}      
\definecolor{colTrajBg}{RGB}{70,80,100}   
\definecolor{colTrajNode}{RGB}{40,40,40}  

\tikzset{
  mybox/.style={
    rectangle, rounded corners=5pt, draw=none,
    text=white, font=\bfseries, align=center,
    minimum width=2.4cm, minimum height=1.5cm
  },
  innerbox/.style={
    mybox, fill=colInner,
    minimum width=3.0cm, minimum height=1.5cm
  },
  condbox/.style={
    mybox, fill=colInner,
    minimum width=3.2cm, minimum height=1.0cm,
    font=\bfseries\large
  },
  trajnode/.style={
    circle, fill=colTrajNode, draw=white, line width=1pt,
    text=white, font=\bfseries\scriptsize,
    minimum size=0.6cm, inner sep=0pt
  },
  arr/.style={
    -{Stealth[length=7pt, width=5pt]},
    line width=1.8pt, color=black!75
  },
}

\pgfplotsset{
  resultplot/.style={
    width=\linewidth,
    height=5.5cm,
    xlabel={Timestep},
    xmin=0,
    xtick distance=20000,
    xticklabel style={font=\scriptsize, /pgf/number format/fixed},
    yticklabel style={font=\scriptsize},
    xlabel style={font=\footnotesize},
    ylabel style={font=\footnotesize},
    title style={font=\footnotesize\bfseries},
    legend style={draw=none, fill=none},
    legend cell align=left,
    grid=both,
    grid style={line width=0.3pt, draw=gray!30},
    tick style={draw=none},
    line width=1.2pt,
    scaled x ticks=false,
    each nth point=10,
    filter discard warning=false,
  }
}

\newcommand{\plotlinewidth}{1.5pt}  

\newcommand{\plotwithband}[5][solid]{%
  \addplot[#2, opacity=0, name path=#3_u, forget plot]
    table[x=timestep, y=#5_upper, col sep=comma]{#4};
  \addplot[#2, opacity=0, name path=#3_l, forget plot]
    table[x=timestep, y=#5_lower, col sep=comma]{#4};
  \addplot[#2, fill opacity=0.2, forget plot]
    fill between[of=#3_u and #3_l];
  \addplot[#2, line width=\plotlinewidth, #1]
    table[x=timestep, y=#5_avg, col sep=comma]{#4};
}

\newcommand{\legendline}[3]{%
  \textcolor{#1}{\tikz[baseline=-0.7ex] \draw[#2, line width=1.5pt] (0,0) -- (1em, 0);}~#3%
}

\newcommand{\algBaseline}{\legendline{clrBaseline}{dashed}{Baseline PPO\cite{schulman2017proximalpo}}}
\newcommand{\algRAC}{\legendline{clrRAC}{dashed}{Reversibility Aware Control\cite{grinsztajn2021there}}}
\newcommand{\algLNT}{\legendline{clrLNT}{dashed}{Leave No Trace\cite{eysenbach2017leave}}}
\newcommand{\algOracle}{\legendline{clrOracle}{dashed}{PPO with Safety Oracle shield\cite{alshiekh2018safe}}}
\newcommand{\algSAVMPC}{\legendline{clrSAVMPC}{solid}{PPO with SAVMPC shield (Ours)}}

\title{\LARGE \bf
Approximating Safety Feedback Without a Safety Oracle\\via Model Predictive Control
}

\author{Jeff Pflueger$^{1}$ and Michael Everett$^{1}$
\thanks{$^{1}$ Northeastern University, Boston, MA, USA
        {\tt\small\{ pflueger.j, m.everett\} @northeastern.edu}}%
}

\begin{document}

\maketitle
\thispagestyle{empty}
\pagestyle{empty}

\begin{abstract}
Safe decision-making algorithms for control of mobile robots often require the existence of feedback to verify the safety of proposed actions. This feedback is assumed to be directly available during the development or deployment of the control system. It can take the form of either an explicit constraint formulation or a set of hand-labeled safety data, both of which can be inaccurate or time consuming to produce. Many recently developed simulators can handle complex interactions and varied environments. These environments have implicit safety constraints that may be hard to model. By leveraging one of these simulators, we can construct a proxy for a safety function that bypasses the need for hand designed feedback in capturing these constraints. We present an algorithm that approximates safety by using reversibility and a positive-invariance assumption on the unsafe state space. This method employs the Model-Predictive Path Integral algorithm (MPPI) to establish this reversibility and verify a proposed action. First the action is projected via the simulator to a future state. Then if MPPI can find a path back to a previous state in the trajectory, that state is guaranteed to be outside the unsafe (positive invariant) set. Experimental results demonstrate that the proposed algorithm can approximate the performance of a safety oracle while avoiding classification of unsafe states as safe.

\end{abstract}

\section{Introduction}

Current methods of ensuring the safety of an agent in decision making problems can provide safety guarantees but encounter a fundamental problem: they typically require prior knowledge of the safety specifications of the agent’s environment. For example, approaches based on control barrier functions\cite{ames_control_2019} or backward reachable tubes\cite{bansal_hamilton-jacobi_2017} reason about states and actions that could violate the known safety constraints in the future. In safe reinforcement learning \cite{brunke2022safe}, discriminator methods~\cite{bharadhwaj2021conservative,srinivasan2020learning,xie2022when} assume access to a safety oracle during the training process and are prone to taking unsafe actions as their discriminators learn. Gaussian process methods\cite{turchetta2016safe,wachi2020safe} require strong assumptions about the underlying safety function (e.g., smoothness) and must be able to observe its true value for feedback. Shielding methods~\cite{alshiekh2018safe,bastani2021safe,Li2019RobustMP} project unsafe actions back to safety. All of these methods require access to ground truth safety information, be it at runtime, or during development to label data. Instead, this paper investigates an alternative method of acquiring safety feedback.

\begin{figure}[t]
\centering
\includegraphics[width=2.5in]{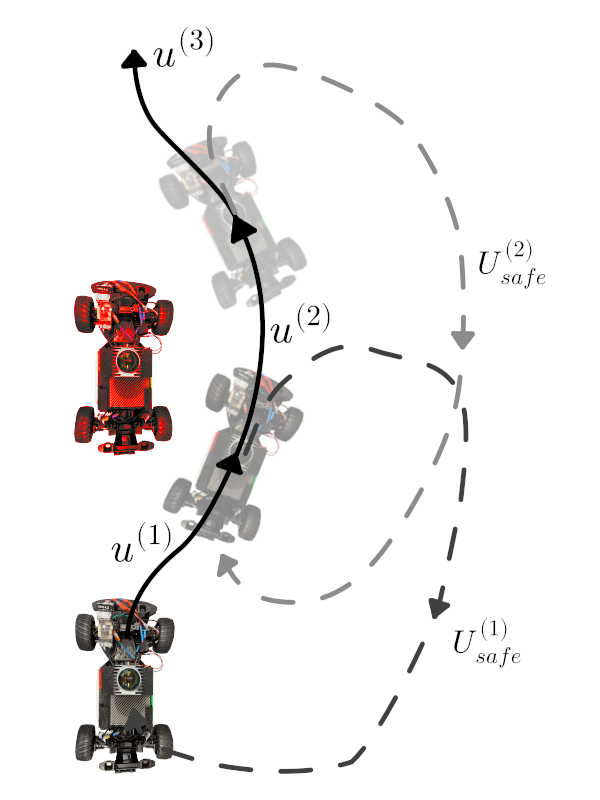}
\caption{Illustration of SAVMPC. The agent is attempting to overtake the vehicle in red. First, SAVMPC projects an action forward. It then ensures there is a set of actions back to the original state before execution.}
\label{fig:SAVMPC}
\end{figure}

In particular, this paper focuses on problems where it is impossible to recover from a violation of the safety constraints, i.e. the unsafe state space is positive invariant. This class of problem divides the state space into $\safe$ where the constraints are satisfied, and $\unsafe$ where at least one is not. While stringent, this assumption yields a clean separation between $\safe$ and $\unsafe$ that makes \textit{reversibility} a well-defined proxy for safety~\cite{kruusmaa2007dont}. A reversibility between two states means that sequences of actions exist that transition an agent between those states, in both directions. Since $\unsafe$ is positive invariant, establishing a reversibility also establishes that both states belong to the same set. Further, establishing a chain of reversibilities to a known safe set also establishes the safety of every state in the reversibility chain. 

To address these challenges, this paper proposes \textbf{S}afety \textbf{A}ssessment \textbf{V}ia \textbf{M}odel-\textbf{P}redictive \textbf{C}ontrol (SAVMPC), a filtering algorithm that uses reversibility as a proxy for unknown safety constraints. SAVMPC employs Model Predictive Control (MPC~\cite{borrelli2017predictive}) to plan through a simulator back to a previous state in the trajectory. In doing this, it automatically verifies that a plan exists that reaches the safe set. If it cannot verify the safety of a planned action, SAVMPC aborts the trajectory in favor of resetting the training environment. \cref{fig:SAVMPC} shows a visualization for this process. The end result is an algorithm that can be dropped in to verify the actions of an unsafe policy while only assuming access to a simulator. It does this while enforcing an agent takes zero unsafe actions in the underlying environment, and without requiring direct access to a safety oracle.

Our primary contributions include:
\begin{itemize}
    \item Safety Assessment Via Model-Predictive Control (SAVMPC), an algorithm that plans back to previous (known safe) states, enabling safe action selection without requiring hand-constructed constraints.
    \item An extension of the proposed reversibility planning algorithm to explicitly calculate a conservative estimate of unknown environment constraints.
    \item Evaluation of SAVMPC as a shield during reinforcement learning training, showing better performance than other reversibility-based approaches, and approximating the results of a method outfitted with an oracle-based shield.
\end{itemize}

\section{Problem Statement}

We model the safe decision making problem as a continuous Constrained Markov Decision Process (CMDP) \cite{altman1999constrained}, using the tuple ($\statespace$, $\actionspace$, $\dyn$, $R$, $\gamma$, $g$). The state space $\statespace \subseteq \mathbb{R}^n$ is divided into two subsets: $\safe$ and $\unsafe$, where $\safe = \mathcal{X} \setminus\unsafe $, and $\unsafe$ is positive invariant. $\actionspace \subseteq \mathbb{R}^m$ is the action space, $\dyn: \mathcal{X} \times \mathcal{U} \to \mathcal{X}$ is the deterministic transition function, $R: \mathcal{X}\to\mathbb{R}$ is the reward function, and $g: \mathcal{X} \to \{0, 1\}$ is the latent safety function. This function returns 1 when $x \in \safe$ and 0 when $x \in \unsafe$ We assume that the dynamics function $\dyn$ is a ``black box'' in that it can be queried but its functional form is unknown. The objective is then to find an approximation $\bar{g}(x)$ such that $\bar{g}(x) = 0,\; \forall x\in \unsafe$. That is, $\bar{g}(x)$ must have no false negatives, though false positives (conservatively labeling safe states as unsafe) are permissible. These specifications reflect practical settings in which engineers can simulate system trajectories and measure success, but safety depends on complex or incompletely understood phenomena that preclude a closed-form constraint

Some examples which embody these specifications:
\begin{itemize}
    \item \textbf{High speed navigation on uneven terrain:} A mobile robot must navigate towards a goal at high speeds on highly varied terrain. The trajectory ends when something causes the robot to break or be unable to proceed. In this environment, latent constraints that determine whether a robot flips or breaks an actuator are unclear, but can be simulated.
    \item \textbf{Robotic manipulation in open work cells:} A stationary manipulator is tasked with inserting pegs into a board in a factory, maximizing speed and success. Moving the manipulator too wildly into collision may damage the end effector, and dropping pegs may leave them outside the work cell, requiring human intervention. 
\end{itemize}
While success and failure are both defined in these environments, the set of constraints on states and actions that lead to a failure are unclear. Our goal is to find a model for the failure set that bypasses the need to know this set of constraints.

\section{Related Work}

\begin{figure*}[b]
    \centering
    \includegraphics[width=0.8\textwidth]{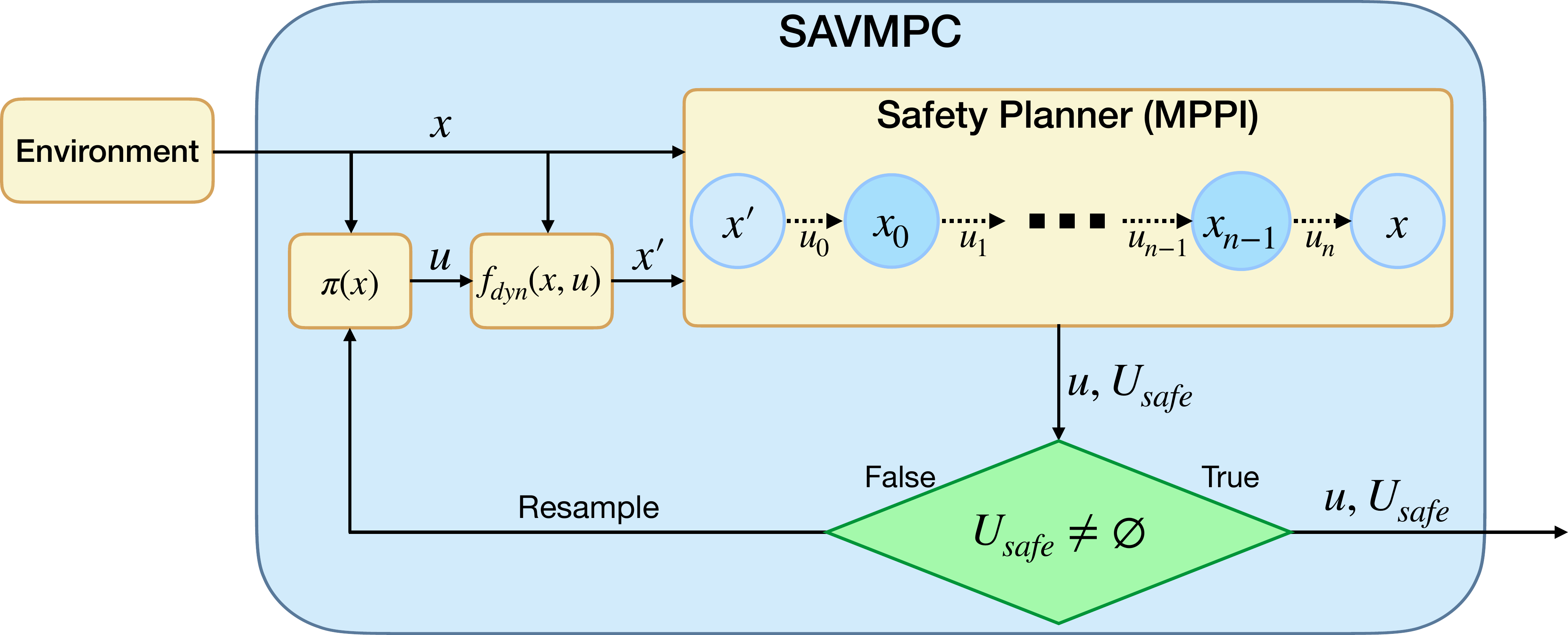}
    \caption{Block diagram of SAVMPC. At state $x$, the algorithm samples an action from a policy $\pi$. It then propagates the action forward to get the induced state $x'$. The safety planner tries to find a set of actions $U_{safe}$ which returns the agent to the previous state. If such a sequence exists, the original action is sent to the agent. Otherwise, SAVMPC resamples from $\pi$.}
    \label{fig:block-diagram}
\end{figure*}

\subsection{Safety Function Approximations in CMDPs}

One way to determine whether an action is safe to implement is to approximate the underlying safety function after taking that action, $g(\dyn(x, u))$. For example, the use of Gaussian Processes (GP) to model safety in discrete MDPs was pioneered in the SafeMDP~\cite{turchetta2016safe} algorithm. This uses a set of uncertainty bounds provided by the GP to plan to new safe locations for exploration. This work also uses the concept of reversibility, enforcing that an agent can return to known safe states from future states that it visits. SNO-MDP~\cite{wachi2020safe} builds on this work with the focus of expanding pessimistic-safe regions for exploitation. While these methods can safely explore MDPs, they are limited to discrete, deterministic MDPs, with Lipschitz assumptions about the underlying safety function. SABRE~\cite{bennett2023provable} takes an active learning approach, seeking to find the best safety function from a class of functions. In this case, SABRE models safety as a set of Generalized Linear Models (GLM), and incentivizes exploration in regions of safety disagreement between set members. LoBiSaRL~\cite{wachi2024long-term} also models safety as a GLM, but combines the approximation with Lipschitz assumptions. Both SABRE and LoBiSaRL are able to explore stochastic MDPs while taking little to no unsafe actions. However, they both rely on a known safe policy. All discussed methods require an observable ground truth safety function to label visited states. The assumptions on the structure of the safety function make it difficult to deploy these algorithms in the real world. 

There are several techniques that leverage deep learning to approximate safety in Reinforcement Learning\cite{sutton1998reinforcement}. SQRL~\cite{srinivasan2020learning} takes a transfer learning approach, learning a Q function as a safety discriminator in a pre-training phase and deploying it to fine-tune a learned policy. Conservative Safety Critic~\cite{bharadhwaj2021conservative} also learns a Q function as a discriminator but leverages Conservative Q Learning~\cite{kumar2020conservative} to over-approximate the true probability of a state being unsafe in expectation. PAINT~\cite{xie2022when} trains a safety discriminator in concert with an reinforcement learning~\cite{sutton1998reinforcement} policy, allowing it to dictate when a training episode ends. Each of these methods will encounter safety violations during exploration, and require some knowledge of a ground truth safety function to label data.
A set of methods try to leverage reversibility to establish safety during training. Leave No Trace~\cite{eysenbach2017leave} takes the approach of learning a reset policy in tandem with the forward policy. The reset policy is employed to both determine when to reset a training episode, and rollback the environment for the next episode. Reversibility Aware Exploration~\cite{grinsztajn2021there} includes a notion of reversibility as well. It simultaneously trains a discriminator to predict the probability of being able to return to a state in a trajectory, and uses that signal to reward a policy for taking reversible actions. Both of these leverage the concept of reversibility as a measure of safety. While these methods learn a proxy for safety without any ground truth representation, they are still prone to violating safety constraints, especially at the beginning of training.

\subsection{Model Predictive Shielding}

Shielding methods are often deployed to enforce that any action sampled from a policy is safe for an agent to take. This concept was codified in Safe Reinforcement Learning via Shielding~\cite{alshiekh2018safe}. This work introduced the concept of a shield in Reinforcement Learning and two rules that must be satisfied: The shield must always be correct with respect to its concept of safety and must minimally interfere with the RL agent. Model Predictive Control based Shielding (MPCS) seeks to use a planner to ensure there is always some backup policy that leads to a known safe region. This was first deployed~\cite{bastani2021safe} in a deterministic MDP on the cartpole environment. MPCS is employed to ensure at each future step that there is a plan back to the region of attraction for a stabilizing LQR controller. An extension into a stochastic environments, Robust Model Predictive Shielding (RMPS) ~\cite{Li2019RobustMP} uses a robust MPC controller and monte-carlo sampling method for establishing probabilistic recovery guarantees. Dynamic Model Predictive Shielding (DMPS)~\cite{banerjee2024dynamic} adds an objective to the model predictive shield, aiming to find the plan back to a safe region of the best reward. While these algorithms can provide safety guarantees for actions, they require a representation of the safe state space on which to plan.

\section{Approach}
\subsection{Safety Definition}

Constrained Markov Decision Processes \cite{altman1999constrained} have a set of constraints which define the safe and unsafe state space. $\safe$ is the set of states that do not violate any of these constraints. $\unsafe$ is the set of states in which at least one constraint is violated. It follows that the safe and unsafe state spaces are complements, i.e. $\unsafe = \mathcal{X} \setminus \safe$. Together with the assumption that $\unsafe$ is positive invariant, finding which subset contains a given state serves as a proxy for safety. By finding a plan from a state $x$ to some state $x' \in \safe$ that does not intersect $\unsafe$, we can verify the safety of $x$. For SAVMPC, this takes the form of a reversible relationship between states. We detail modeling definitions and discuss how we establish reversibility as a safety proxy below.

\begin{definition}[Trajectory] A \textbf{trajectory} $\mathcal{T}=\{(x_i, u_i)\ |\ x_{i+1} = f(x_i, u_i)\ \forall i \in [0,..., H-1]\}$ is a sequence of state-action pairs, with time duration $H$.

\end{definition}

\begin{definition}[Reversibility]
     A \textbf{reversibility} between two states $x$ and $x'$ is a sequence of actions that transitions an agent from state $x$ to $x'$, and vice-versa.
\end{definition}

\begin{definition}[Safety]
    A state $x\in\mathcal{X}$ is called \textbf{safe} if there exists a reversibility from $x$ to some state $x'\in\safe$. A trajectory is called \textbf{safe} if there exists such a reversibility for every state in the trajectory. 
\end{definition}

This safety definition enables the iterative verification of an entire trajectory. By starting from the safe state space $\safe$, and finding that a reversibility exists for the result of every proposed action, an algorithm can verify the safety of a trajectory. Furthermore, this definition does not rely on prior knowledge of safety constraints on the system. As long as we can show at every timestep there is a plan back to the previous state, then the current state is safe.

\subsection{Safety Approximation with MPPI}

\begin{figure*}[b]
  \centering

  \begin{minipage}[t]{0.3\linewidth}
    \centering
    \includegraphics[width=\linewidth]{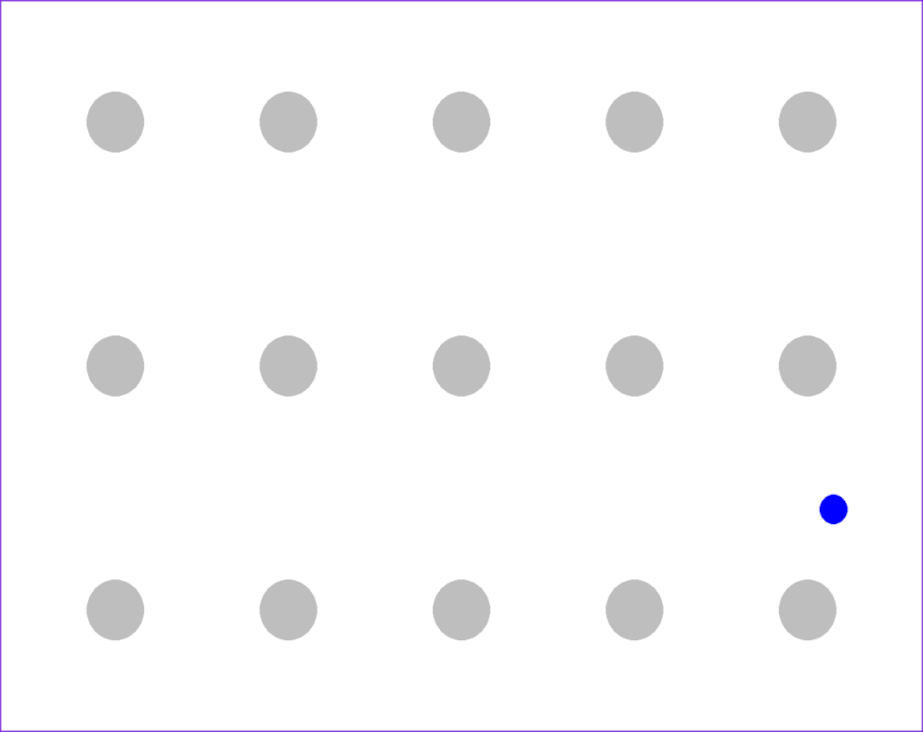}
    \subcaption{Environment}
    \label{fig:obstacle-env}
  \end{minipage}
  \hfill
  \begin{minipage}[t]{0.3\linewidth}
    \centering
    \includegraphics[width=\linewidth]{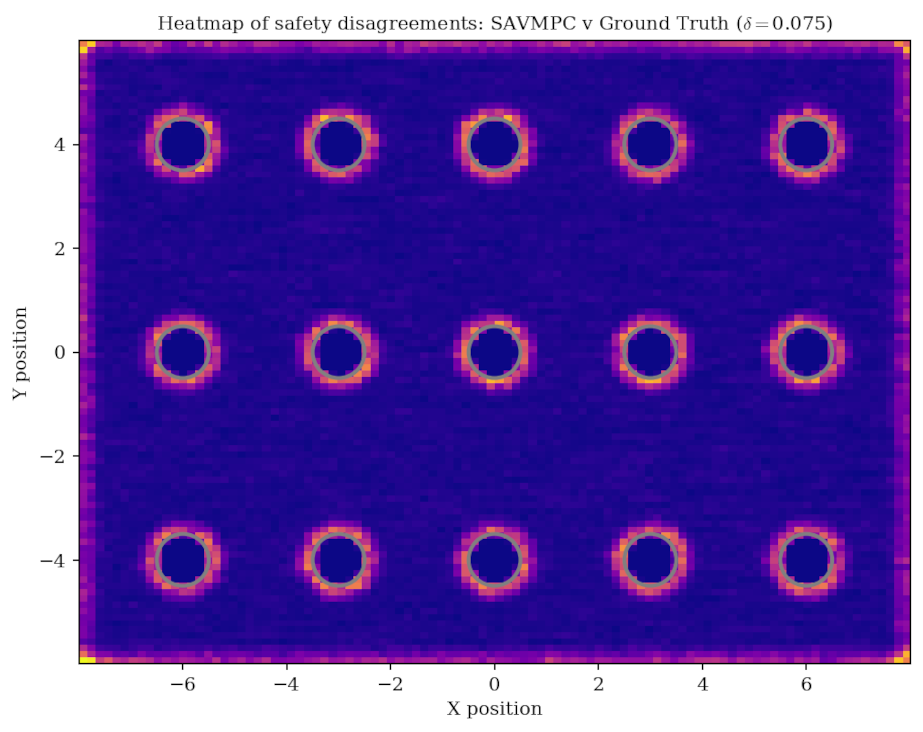}
    \subcaption{Heatmap: $\delta=0.075$}
    \label{fig:conservative-heatmap}
  \end{minipage}
  \hfill
  \begin{minipage}[t]{0.3\linewidth}
    \centering
    \includegraphics[width=\linewidth]{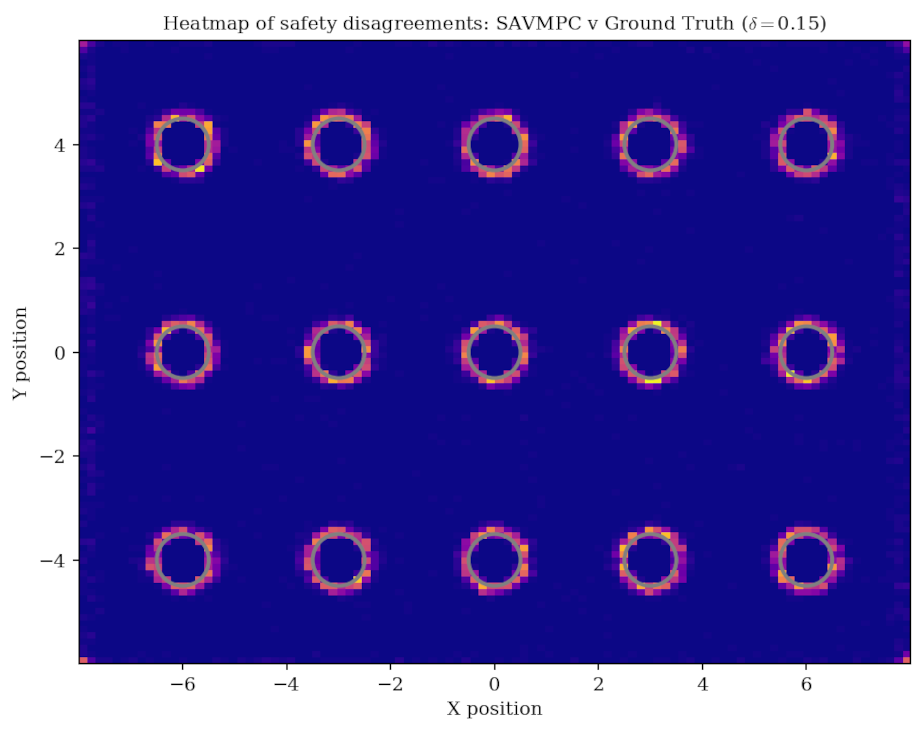}
    \subcaption{Heatmap: $\delta=0.15$}
    \label{fig:loose-heatmap}
  \end{minipage}

  \caption{Set of images depicting how SAVMPC differs with the ground truth safety in an environment with obstacles. \cref{fig:obstacle-env} shows the operating environment. An agent (blue dot) acts according to a unicycle model, controlled by a random policy. Any collision with the obstacles (grey circles) will terminate the agent's trajectory. \cref{fig:loose-heatmap} and \cref{fig:conservative-heatmap} depict heatmaps of where SAVMPC disagrees with the ground truth safety function.}
  
  \label{fig:heatmaps}
\end{figure*}
With safety defined, finding a reversibility between each state in a trajectory and the next state will verify the safety of the trajectory. This can be modeled as a feasibility problem, which aims to find any set of actions that leads from starting state $x'$ through goal state $x$:
\begin{equation}
    \label{eq:reversibility-opt}
    \begin{aligned}
        \textbf{find} \quad &u_0, u_1, \ldots \in \mathcal{U} \\
        \text{subject to} \quad & x_0 = x' \\
        & x_{k+1} = \dyn(x_k, u_k), \quad k \geq 0 \\
        & \exists \, k \geq 0 : x_k = x
    \end{aligned}
\end{equation}

A feasible solution to \cref{eq:reversibility-opt} serves as a safety certificate; however, the black-box transition dynamics make it difficult to solve the problem exactly.
Instead, this paper proposes a sampling-based approach to solve a mildly relaxed version of \cref{eq:reversibility-opt}:
\begin{equation}
    \label{eq:reversibility-opt-relaxed}
    \begin{aligned}
        \textbf{find} \quad & u_0, \ldots, u_{T-1} \in \mathcal{U} \\
        \text{subject to} \quad & x_0 = x' \\
        & x_{k+1} = f_{\text{dyn}}(x_k, u_k), \quad k \in \{0, \ldots, T-1\} \\
        & \exists \, k \in \{0, \ldots, T\} : \|x_k - x\|_2 \leq \delta,
    \end{aligned}
\end{equation}

 which introduces a time horizon $T$ and a safety/conservativeness tolerance $\delta$ to the goal state. These approximations introduce a tradeoff between the computational efficiency of the sampling-based planner and the approximation gap between \cref{eq:reversibility-opt} and \cref{eq:reversibility-opt-relaxed}. The output set of control sequences that verify the trajectory act as a backup policy for the agent. In the case where \cref{eq:reversibility-opt-relaxed} does not provide a solution, the agent can return to its initial state by following these sequences in reverse. 
 
The SAVMPC algorithm for safe action selection is shown in \cref{alg:safety-function}, which assesses the reversibility of states to approximate the true safety function $g(x)$. At a high level, the SAVMPC check takes an action from the agent, propagates it forward in time to get the induced next state, then tries to find a sequence of actions back to the original state to verify safety. To do this, SAVMPC uses the Model Predictive Path Integral (MPPI)~\cite{williams2016aggressive} algorithm for its ability to use a black-box transition function. It can also be parallelized for environments which require many safety checks per second. If MPPI finds a sequence of safe actions back to the original state, then the proposed action is implemented. If it does not, the action is aborted and is resampled from the control policy. After $N$ resamples, the algorithm terminates and the trajectory is aborted.

\begin{algorithm}[b]
\caption{SAVMPC: Safe Action Selection}
\label{alg:safety-function}
\begin{algorithmic}[1]
\Input $x \gets$ State , $\pi \gets$ Control Policy, $\dyn \gets $ Transition Dynamics,
\Require $N_{\text{samples}} \gets$ Maximum samples, $T \gets$ Time horizon, $\delta \gets$ safety tolerance
\Output Control action $u$, Safe abort trajectory $U_{safe}$
\Function{SAVMPC}{$x, \pi, \dyn$}
\For{sample $n = 1, \ldots, N_{\text{samples}}$}
    \State $u \sim \pi(x)$
    \State  $x' \leftarrow \dyn(x, u)$
    \State $U_{\text{safe}} \leftarrow \textbf{MPPI}(x', x, f_{\text{dyn}}, T, \delta)$
    
    \If{$U_{\text{safe}} \neq \emptyset$}
        \State \Return $(u, U_{\text{safe}})$ 
    \EndIf
\EndFor

\State \Return $(u, \emptyset)$
\EndFunction
\end{algorithmic}
\end{algorithm}

\section{Experiments}

\subsection{Disagreement Study}
To look at the quality of the approximation SAVMPC makes, we analyze the regions where it disagrees with the ground truth safety function of an environment with several safety constraints. This environment is composed of a robotic agent, fifteen evenly spaced obstacles, and a collision boundary around the operating space. The agent operates according to a kinematic unicycle model. The obstacles and collision boundary terminate the agent's trajectory if it gets too close, acting as a set of safety constraints. To look at the quality of SAVMPC's approximation, we run a simulation of 300,000 trajectories for 10 timesteps each, governed by a random policy. SAVMPC is run at each timestep to verify the safety of the proposed action. Then the result is compared to the output from a safety oracle. Any place where the outputs of SAVMPC and the oracle differ is referred to as a \textit{disagreement}. We run this experiment for two different safety tolerances. We analyze these experiments in \cref{tab:disagreements}. Here, both experiments are measured in terms of total disagreements, and the average distance from a disagreement to a constraint. The conservative experiment ($\delta=0.075$) yields an order of magnitude more disagreements, with a much higher average distance to the constraints. Disagreements are plotted in \cref{fig:heatmaps} as a set of heatmaps.
\begin{table}[b]
  \centering
  \small
  \begin{tabular}{lcc}
    \toprule
    \textbf{$\delta$}
      & \textbf{Disagreements}
      & \textbf{Avg. Distance to Constraint} \\
    \midrule
    $0.075$ & $264{,}064$ & $0.302$ \\
    $0.150$ & $27{,}974$  & $0.081$ \\
    \bottomrule
  \end{tabular}
  \caption{Experiments measuring SAVMPC v ground truth safety disagreements. Experiments are conducted at two different safety tolerances. Trends show that lowering the safety tolerance increases how conservative SAVMPC is with regard to approximating ground truth safety.}
  \label{tab:disagreements}
\end{table}
The conservative policy heatmap is shown in \cref{fig:conservative-heatmap}. The small tolerance means it is more difficult for SAVMPC to find a reversibility, particularly around the constraint boundaries. This is consistent with the heatmap. The disagreements are clearly grouped around the obstacles and boundaries, with hot spots as they get closer to them. There are also disagreements spread throughout the map. The less conservative policy is shown in \cref{fig:loose-heatmap}. Here, the disagreements are less prominent, though they still exist. Disagreements are closer to the obstacles, with significantly fewer at the boundaries of the map. There are also far fewer disagreements in the space between obstacles. Grey circles in the heatmaps indicate where the obstacle boundaries are placed. The hotspots on both maps are concentrated around these circles, but never within. Both SAVMPC parameterizations achieved no false negative safety detections throughout both experiments. These experiments show that SAVMPC is able to over-approximate the true boundary of the safety constraints, yielding a conservative estimate of the ground truth. The safety tolerance serves as a tunable parameter, adjusting how close of an approximation to make, with the tradeoff that a looser tolerance is more likely to produce false negative disagreements.

\begin{figure*}[t]
  \centering
  \pgfplotsset{table/search path={anc}}

  \begin{subfigure}[t]{0.48\textwidth}
    \begin{tikzpicture}
      \begin{axis}[
        resultplot,
        title={Training Reward (Higher is Better)},
        ylabel={Reward},
        ymin=-20,
        xmax=100000,
        xtick={0,20000,40000,60000,80000},
      ]
        \plotwithband[dashed]{clrBaseline}{blr}{cc_baseline_total_reward.csv}     {total_reward}
        \plotwithband[dashed]{clrRAC}     {rcr}{cc_rac_total_reward.csv}          {total_reward}
        \plotwithband[dashed]{clrLNT}     {lnr}{cc_lnt_total_reward.csv}          {total_reward}
        \plotwithband[dashed]{clrOracle}  {orr}{cc_fully_accurate_total_reward.csv}{total_reward}
        \plotwithband{clrSAVMPC}  {svr}{cc_rl-savmpc_total_reward.csv}    {total_reward}
      \end{axis}
    \end{tikzpicture}
    \subcaption{Training reward over timesteps.}
    \label{fig:cartpole-reward}
  \end{subfigure}
  \hfill
  \begin{subfigure}[t]{0.48\textwidth}
    \begin{tikzpicture}
      \begin{axis}[
        resultplot,
        title={Trajectory Aborts (Lower is Better)},
        ylabel={Percent of episodes aborted},
        ymin=0, ymax=110,
        xmax=100000,
        ytick={0,20,40,60,80,100},
        xtick={0,20000,40000,60000,80000},
      ]
        \plotwithband[dashed]{clrBaseline}{bls}{cc_baseline_is_stuck.csv}         {is_stuck}
        \plotwithband[dashed]{clrRAC}     {rcs}{cc_rac_is_stuck.csv}              {is_stuck}
        \plotwithband[dashed]{clrLNT}     {lns}{cc_lnt_is_stuck.csv}              {is_stuck}
        \plotwithband[dashed]{clrOracle}  {ors}{cc_fully_accurate_is_stuck.csv}   {is_stuck}
        \plotwithband{clrSAVMPC}  {svs}{cc_rl-savmpc_is_stuck.csv}        {is_stuck}
      \end{axis}
    \end{tikzpicture}
    \subcaption{Trajectory aborts over timesteps.}
    \label{fig:cartpole-aborts}
  \end{subfigure}

  \vspace{0.8em}

  \begin{subfigure}[t]{\textwidth}
  \centering
  \small
  \begin{tabular}{lccc}
    \toprule
    \textbf{Algorithm}
      & \textbf{Final Reward (Avg)}
      & \textbf{Constraint Violations (\% of episodes)}
      & \textbf{Usable?} \\
    \midrule
    \algBaseline & $542.38 \pm 40.43$  & $77.3\% \pm 1.4\%$  & No: Constraints \\
    \algRAC & $539.15 \pm 18.04$  & $35.1\% \pm 5.9\%$  & No: Constraints \\
    \algLNT & $119.56 \pm 4.39$   & $58.5\% \pm 11.8\%$ & No: Constraints \\
    \algOracle & $537.76 \pm 21.96$  & $0.0\% \pm 0.0\%$   & No: Oracle \\
    \algSAVMPC & $533.66 \pm 32.14$  & $0.0\% \pm 0.0\%$   & \textbf{Yes} \\
    \bottomrule
  \end{tabular}
  \subcaption{Summary of final performance metrics.}
  \label{tab:cartpole-results}
  \end{subfigure}

  \caption{Results from Continuous Cartpole environment. Training reward and
           trajectory aborts across algorithms are averaged over 10 seeds. Shaded regions denote $\pm 1$ standard deviation. Data smoothed via rolling average with a Savitzky-Golay filter. SAVMPC matches the performance of the safety oracle while maintaining zero constraint violations over training.}
  \label{fig:results-continuous-cartpole}
\end{figure*}

\subsection{RL Experiments}

We present experiments in the safe reinforcement learning domain. Experiments are composed of training five different agents across two separate environments.
\begin{itemize}
    \item {\textbf{PPO~\cite{schulman2017proximalpo}:}} An unmodified PPO algorithm from the stable-baselines~\cite{stable-baselines3} framework that does not consider safety.
    \item {\textbf{Reversibility Aware Control (RAC)~\cite{grinsztajn2021there}:}} A PPO algorithm that approximates the reversibility of an action by learning a reversibility estimator during training. If a state-action pair does not meet a reversibility threshold, it is rejected, and the action is resampled. After achieving a maximum number of resamples, the algorithm will terminate the current trajectory.
    \item {\textbf{Leave No Trace (LNT)~\cite{eysenbach2017leave}:}} An SAC algorithm which simultaneously trains a forward policy, and reverse policy to reset the system. LNT switches between policies if a minimum reverse Q value is not met. 
    \item {\textbf{PPO with Safety Oracle ~\cite{alshiekh2018safe}:}} A PPO algorithm outfitted with a safety oracle based shield. This shield can query the oracle on future states and determine when an action would lead to an unsafe state. If an unsafe state is predicted, it will trigger resampling an action from the policy. After achieving a maximum number of resamples, it will terminate the current trajectory. This comparison is meant to illustrate the difference between the true safety function, and SAVMPC's approximation of it.
    \item {\textbf{PPO with SAVMPC:}} A PPO algorithm outfitted with SAVMPC employed as a shield. This shield uses the SAVMPC algorithm to establish reversibility to the previous state in the trajectory. If a reversibility can’t be established, the shield will trigger the agent resampling from the RL policy. After achieving a maximum number of resamples, the shield will terminate the current trajectory.
\end{itemize}
The shielding method with the safety oracle is the closest point of comparison to SAVMPC. It directly implements the function that SAVMPC seeks to approximate. We also compare against RAC and LNT, both of which are methods that operate in the same domain, i.e. safe RL without direct access to a safety oracle. 

We present results across training in two different environments: Continuous Cartpole and Two Dimensional Navigation. Both of these environments are designed to drive an agent towards safety constraints that terminate the agent's trajectory. The purpose of this is to showcase the safety properties of the implemented algorithms.
\subsubsection{\textbf{Continuous Cartpole Environment}}

Continuous Cartpole is a modified cartpole environment, adapted from the gymnasium~\cite{towers2024gymnasium} implementation. It features a continuous action space, as well as a modified reward function which incentivizes approaching the state constraints. The state is represented by four values, $x=(p_x, v_x, \theta, \omega)$ where $p_x$ is the horizontal position of the cart, $v_x$ is the horizontal velocity of the cart, $\theta$ is the angle of the pole, and $\omega$ is the angular velocity of the pole. The agent controls the system by pushing the cart left or right. The angle of the pole is constrained to be within $12^{\circ}$ of vertical. The reward function is defined in \cref{eq:cartpole-rew}.

\begin{equation}
\label{eq:cartpole-rew}
    R(x) = 1 + \lambda|\theta|
\end{equation}

 The goal of the reward is to encourage the agent to approach the constraints of the pole without violating them, rewarding riskier, but safe behaviors. The inclusion of $\theta$ as a reward parameter incentivizes the agent to act closer to the state constraints. $\lambda$ is a scaling hyperparameter which determines how much to reward the agent for an aggressive policy. A training episode ends if the pole angle exceeds its constraints, or if a time limit is reached.

In \cref{fig:results-continuous-cartpole} we visualize the experimental results for the cartpole environment. The agent outfitted with the SAVMPC shield is able to match the performance of the PPO algorithm, and the oracle based shield. It also outperforms both reversibility-aware algorithms initially. We attribute this to directly measuring reversibility rather than having to learn an approximation for it. As RAC learns its discriminator it is able to match the performance of the SAVMPC shield, but still encounters constraint violations along the way. The SAVMPC equipped shield is better able to explore the state space early due to having a complete approximation. The SAVMPC shield matches the performance of the oracle in trajectory aborts. In combination with matching the performance of the oracle in reward, these results suggest that SAVMPC closely approximates the safety oracle. In the cartpole environment, the agent constantly needs to adjust to avoid the pole falling and violating constraints. The agent equipped with SAVMPC converges faster than the other agents to a final reward, because its conservative approximation shrinks the exploration space. SAVMPC also settles to a steady number of trajectory aborts slightly faster than the oracle based shield. They both converge to similar abort frequencies as training continues. The similar end values of both agents further provide evidence of SAVMPC as an approximation of the safety oracle.

\begin{figure*}[t]
  \centering
  \pgfplotsset{table/search path={anc}}

  \begin{subfigure}[t]{0.48\textwidth}
    \begin{tikzpicture}
      \begin{axis}[
        resultplot,
        title={Training Reward (Higher is Better)},
        ylabel={Reward},
        ymin=-200,
        xmax=100000,
        xtick={0,20000,40000,60000,80000},
      ]
        \plotwithband[dashed]{clrBaseline}{blr}{td_baseline_total_reward.csv}     {total_reward}
        \plotwithband[dashed]{clrRAC}     {rcr}{td_rac_total_reward.csv}          {total_reward}
        \plotwithband[dashed]{clrLNT}     {lnr}{td_lnt_total_reward.csv}          {total_reward}
        \plotwithband[dashed]{clrOracle}  {orr}{td_fully_accurate_total_reward.csv}{total_reward}
        \plotwithband{clrSAVMPC}  {svr}{td_rl-savmpc_total_reward.csv}    {total_reward}
      \end{axis}
    \end{tikzpicture}
    \subcaption{Training reward over timesteps.}
    \label{fig:twod-nav-reward}
  \end{subfigure}
  \hfill
  \begin{subfigure}[t]{0.48\textwidth}
    \begin{tikzpicture}
      \begin{axis}[
        resultplot,
        title={Trajectory Aborts (Lower is Better)},
        ylabel={Percent of episodes aborted},
        ymin=0, ymax=110,
        xmax=100000,
        ytick={0,20,40,60,80,100},
        xtick={0,20000,40000,60000,80000},
      ]
        \plotwithband[dashed]{clrBaseline}{bls}{td_baseline_is_stuck.csv}         {is_stuck}
        \plotwithband[dashed]{clrRAC}     {rcs}{td_rac_is_stuck.csv}              {is_stuck}
        \plotwithband[dashed]{clrLNT}     {lns}{td_lnt_is_stuck.csv}              {is_stuck}
        \plotwithband[dashed]{clrOracle}  {ors}{td_fully_accurate_is_stuck.csv}   {is_stuck}
        \plotwithband{clrSAVMPC}  {svs}{td_rl-savmpc_is_stuck.csv}        {is_stuck}
      \end{axis}
    \end{tikzpicture}
    \subcaption{Trajectory aborts over timesteps.}
    \label{fig:twod-nav-aborts}
  \end{subfigure}

  \vspace{0.8em}
  \begin{subfigure}[t]{\textwidth}
  \centering
  \small
  \begin{tabular}{lccc}
    \toprule
    \textbf{Algorithm}
      & \textbf{Final Reward (Avg)}
      & \textbf{Constraint Violations (\% of episodes)}
      & \textbf{Usable?} \\
    \midrule
    \algBaseline & $959.44 \pm 49.01$  & $4.6\% \pm 0.9\%$   & No: Constraints \\
    \algRAC & $967.99 \pm 39.54$  & $4.0\% \pm 0.9\%$   & No: Constraints \\
    \algLNT & $375.94 \pm 109.65$ & $6.1\% \pm 3.1\%$   & No: Constraints \\
    \algOracle & $782.23 \pm 88.44$  & $0.0\% \pm 0.0\%$   & No: Oracle \\
    \algSAVMPC & $842.74 \pm 101.16$ & $0.0\% \pm 0.0\%$   & \textbf{Yes} \\
    \bottomrule
  \end{tabular}
  \subcaption{Summary of final performance metrics.}
  \label{tab:twod-nav-obs-results}
  \end{subfigure}

  \caption{Results from Two Dimensional Navigation environment. Training reward and
           trajectory aborts across algorithms over 10 seeds. Shaded regions denote
           $\pm 1$ standard deviation. Data smoothed via rolling average with
           a Savitzky-Golay filter. SAVMPC outperforms the oracle based shield while maintaining zero constraint violations across training}
  \label{fig:results-twod-nav}
\end{figure*}

\subsubsection{\textbf{Two Dimensional Navigation Environment}}

In this environment, the agent tries to find a goal location while avoiding a sinkhole in the middle of the operating region. The state is represented by five values, $x = (p_x, p_y,p_\theta, p_{goalx}, p_{goaly})$, where $p_x$ and $p_y$ represent the agent's position in two dimensions and $p_\theta$ represents its orientation. $p_{goalx}$ and $p_{goaly}$ represent the position of the goal in two dimensions. Both agent and goal positions are randomly spawned at different sides of the environment at the beginning of a training episode, with the sinkhole stationary between them. The agent's position is governed by a kinematic unicycle model, with the control action being a linear and angular velocity command. The agent seeks to find actions to drive it to the goal while avoiding the sinkhole and boundaries of the environment. The reward function is defined in \cref{eq:two-dim-nav-reward}.
\begin{equation}
\label{eq:two-dim-nav-reward}
    R=-d_{goal} + \mathbf{1}_{goal}(x)(1000) + \mathbf{1}_{term}(-100)
\end{equation}
$d_{goal}$ is the euclidean distance to the goal, $\mathbf{1}_{goal}(x)$ is a function indicating whether the agent has reached the goal, and $\mathbf{1}_{term}$ is a function indicating whether the trajectory has been terminated prematurely.
The agent is penalized at each timestep, penalized for an early termination from constraint violations, and rewarded heavily for finding the goal. This reward is structured to incentivize quick navigation, while driving the agent toward the sinkhole between it and the goal. A training episode terminates if the agent leaves the operating region, comes within a set distance of the center of the sinkhole, or if the time limit is reached.

We visualize performance and abort data in \cref{fig:results-twod-nav}. Similarly to the Cartpole environment, SAVMPC initially outperforms both the RAC and LNT algorithms and matches the Baseline PPO in reward. As training continues, it converges to a similar final reward to the oracle based shield. We believe this is also due to SAVMPC conservatively shrinking the exploration space. Because SAVMPC is making a more conservative approximation of the boundaries of the space and the sinkhole, it is able to learn faster than the other algorithms, which either do not shrink the space as much, or have to learn safety approximations over time. This is corroborated by the trajectory abort plot. SAVMPC is aborting much more frequently early in training as it encounters its constraint approximations. As training goes on, it begins to abort less than the safety oracle. Towards the end of training, SAVMPC and the safety oracle converge to similar values. Even though RAC eventually outperforms SAVMPC, like LNT it crucially still encounters constraint violations during training, which are potentially harmful to the agent. SAVMPC is the only agent which converges to a high final reward, while having no constraint violations or privileged information.

\section{Conclusion}

This work introduced SAVMPC, a safety verification algorithm that ensures that any action about to be taken by a robot can be un-done. Instead of relying on detailed knowledge of the safety specifications like much of the existing work, SAVMPC instead uses a proxy that only relies on black box access to the dynamics, and a positive invariance assumption on the unsafe state space. This approach, based on Model Predictive Path Integral (MPPI) control, plans a trajectory from a future state back to the current state to assess whether the system would remain safe under a candidate action. Experimental results highlight SAVMPC as an approximation of ground truth safety information, and demonstrates its performance when acting as a safety shield in reinforcement learning.

\bibliographystyle{IEEEtran}
\bibliography{citations}{}

\end{document}